\documentclass[USletter, 10 pt, conference]{ieeeconf}  
\IEEEoverridecommandlockouts
\usepackage{cite}
\usepackage{amsmath,amssymb,amsfonts}
\usepackage{graphicx}
\usepackage{subcaption}
\usepackage{textcomp}
\usepackage{tcolorbox}
\usepackage{booktabs}
\usepackage{makecell} 
\usepackage{overpic}
\usepackage{siunitx}

\usepackage[linesnumbered]{algorithm2e}
\makeatletter
\newcommand{\nosemic}{\renewcommand{\@endalgocfline}{\relax}}
\newcommand{\dosemic}{\renewcommand{\@endalgocfline}{\algocf@endline}}
\let\oldnl\nl
\newcommand{\nonl}{\renewcommand{\nl}{\let\nl\oldnl}}
\usepackage{romannum}

\usepackage{float}  
\usepackage{hyperref} 
\let\labelindent\relax
\usepackage{enumitem}
\usepackage{threeparttable}

\usepackage{pgfplots}
\usepackage{pdfpages}
\pgfplotsset{compat=1.18}
\usepackage{xcolor}
\usepackage{colortbl}
\usepackage{lpic}

\usepackage{multirow}
\usepackage{array}
\usepackage{colortbl}
\usepackage{hhline}

\IEEEoverridecommandlockouts                              
\title{\LARGE \bf Federated Learning for Data-Driven Feedforward Control: A Case Study on Vehicle Lateral Dynamics}

\author{Jakob Weber$^{1}$, Markus Gurtner$^{1}$, Benedikt Alt$^2$, Adrian Trachte$^{2}$ and Andreas Kugi$^{3}$
\thanks{$^{1}$ Jakob Weber and Markus Gurtner are with the Center for Vision, Automation \& Control, AIT Austrian Institute of Technology GmbH, Vienna, Austria. 
               {\tt\small jakob.weber@ait.ac.at}}
\thanks{$^{2}$ Benedikt Alt and Adrian Trachte are with the Robert Bosch GmbH, Renningen, Germany.}
\thanks{$^{3}$ Andreas Kugi is with the Automation and Control Institute, TU Wien, Austria and the AIT Austrian Institute of Technology GmbH, Vienna, Austria.}
}

\begin{document}

\maketitle
\thispagestyle{empty}
\pagestyle{empty}

\begin{abstract}
In many control systems, tracking accuracy can be enhanced by combining (data-driven) feedforward (FF) control with feedback (FB) control. 
However, designing effective data-driven FF controllers typically requires large amounts of high-quality data and a dedicated design-of-experiment process. 
In practice, relevant data are often distributed across multiple systems, which not only introduces technical challenges but also raises regulatory and privacy concerns regarding data transfer. 
To address these challenges, we propose a framework that integrates Federated Learning (FL) into the data-driven FF control design. 
Each client trains a data-driven, neural FF controller using local data and provides only model updates to the global aggregation process, avoiding the exchange of raw data.
We demonstrate our method through simulation for a vehicle trajectory-tracking task.
Therein, a neural FF controller is learned collaboratively using FL. 
Our results show that the FL-based neural FF controller matches the performance of the centralized neural FF controller while reducing communication overhead and increasing data privacy. 
\end{abstract}

\section{Introduction}
The increasing connectivity in modern control applications has led to an unprecedented availability of data.
Obviously, this data can be leveraged to learn control strategies.
However, three challenges arise when dealing with distributed data in learning scenarios:
\begin{itemize}
    \item \textbf{Data privacy}: regulations and security concerns often prevent sharing raw data across devices. 
    \item \textbf{Communication efficiency}: bandwidth is limited, particularly in large-scale networks or remote situations.
    \item \textbf{System heterogeneity}: devices differ in dynamics, computational resources, and operating as well as environmental conditions.
\end{itemize}
To address these challenges, Federated Learning (FL) has recently emerged as a promising paradigm~\cite{mcmahan2017communication}. 
Instead of transmitting raw data, devices exchange learned models with a central aggregator.
This reduces communication overhead and improves privacy aspects. 
While FL has been extensively explored in the context of perception and prediction~\cite{mcmahan2017communication}, its potential for control applications remains largely untapped.

In this work, we propose a Federated Learning framework for neural feedforward (FF) control. 
By integrating FL with neural FF controllers, we enable connected devices to learn collaboratively without sharing raw data.
Simultaneously, we can improve communication scalability by reducing the amount of transferred data.
The main contribution of this paper is the introduction and simulation-based verification of FL-based Neural FF Control. 
This approach enables communication-efficient, privacy-aware training of neural FF controllers across distributed devices and is demonstrated in a vehicle dynamics scenario. 
The remainder of this paper is structured as follows:
Section~\ref{sec:Related-work} reviews related work on Federated Learning and data-driven feedforward control. 
Section~\ref{sec:FL-and-FF-control} presents details on the proposed framework. 
Section~\ref{sec:Experimental-design} describes the experimental setup, and Section~\ref{sec:Results-and-discussion} discusses the results. 
Finally, Section~\ref{sec:Conclusion-and-outlook} concludes with a summary and outlook.

\section{Related Works} \label{sec:Related-work}
Feedforward control (FF) is frequently applied to reduce the workload of a feedback controller (FB).  
This typically improves tracking performance and efficiency while reducing stability-related issues. 
The two most common FF control concepts are model-based and data-driven.
Model-based FF control is built on system-specific knowledge and ranges from a stationary input-to-output relationship to the inverse system dynamics, e.g.,~\cite{kapania2015_FB_FF_steering_control}.
As a model-free alternative, data-driven FF control aims to learn the inverse stationary or dynamic relationship, see, e.g.,~\cite{zhou2017_NN_FF_trajectory_tracking, aarnoudse2024_NN_FF_Control, tian2022_parallel_learning_steering_control}, and some applications date back to~\cite{hunt1992_NN_for_control}. 

With the rise of connectivity, the research in safe and efficient multi-agent systems is expanding rapidly~\cite{jing2021_learning_controller_multi-agent_systems}. 
New issues, such as communication efficiency, arise because a high-bandwidth connection cannot be guaranteed~\cite{tallat2023industry50_FL}.
Furthermore, when performing control tasks with fast dynamics, sending the necessary data to each client at millisecond sampling rates is challenging.
Additionally, there is a policy-driven momentum for privacy preservation; see~\cite{woisetschläger2024federated}. 

The state-of-the-art method for communication-efficient, privacy-aware, and distributed learning is \textit{Federated Learning} (FL), see~\cite{mcmahan2017communication, kairouz2021advances}. 
Each participating client receives a model from a central server and incorporates its local information by updating the model on its private data. 
The central server then aggregates the learned models (or gradient-like information) in a privacy-aware way.
This results in a global model update, which is then re-distributed to the client.
This process continues for multiple rounds and is described in details in, e.g.,~\cite{mcmahan2017communication, kairouz2021advances}.
Applying FL to control systems is an open topic in the control and machine learning community, see~\cite{weber2024combining_FL_control} for a survey on FL and control. 
Nevertheless, FL has already been applied to enable privacy-aware training of inverse dynamic models for robotic arms, see~\cite{jimenez2025-informed-FL-inverse-dynamics}.

\section{FL-based neural FF control} \label{sec:FL-and-FF-control}
\begin{figure}
    \centering
    \includegraphics[width=0.99\linewidth]{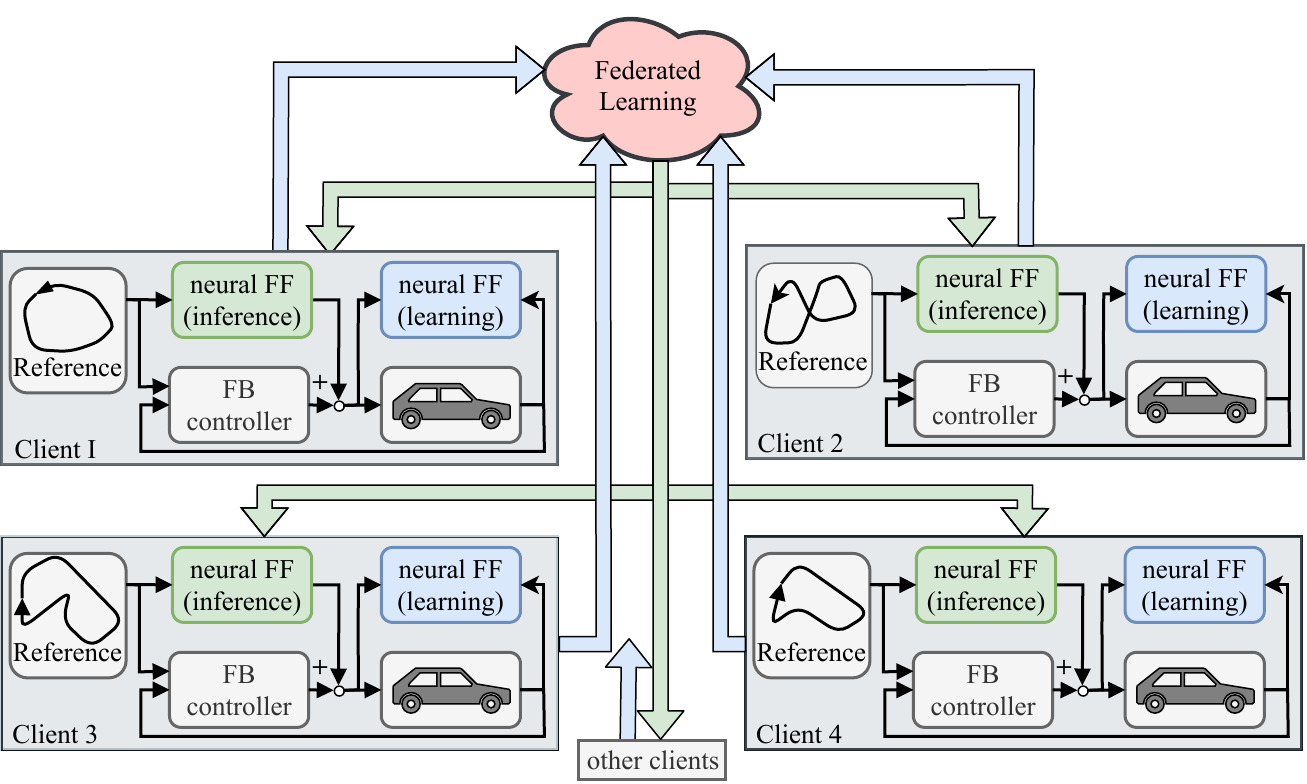}
    \caption{FL-based neural feedforward controller setup for four or more autonomous vehicle clients with individual reference trajectories.
    Each client is equipped with a suitable feedback controller (FB). 
    The neural FF (green) is learned collaboratively using Federated Learning.
    }
    \label{fig:fl-based-FF_sketch}
\end{figure}

Learning a feedforward (FF) controller using neural networks $f_{\theta}$, referred to as a neural FF controller, has become state-of-the-art in modern control systems, see, e.g.,~\cite{hunt1992_NN_for_control, shi2019_neural_lander, zhou2017_NN_FF_trajectory_tracking}. 
Nevertheless, a careful design of experiment (DoE) is necessary to ensure that the neural FF conroller learns in all required operational domains of the control system. 
The time-consuming collection of training data according to the designated DoE can be distributed across multiple similar systems.
This introduces challenges concerning data sharing, such as communication costs and privacy regulations. 
To address these challenges, we propose integrating Federated Learning (FL) with neural FF control.
This enables privacy-aware, communication-efficient, and continuous learning on the control task across multiple clients.
A sketch of the proposed FL-based neural FF controller architecture is given in Fig.~\ref{fig:fl-based-FF_sketch}. 
Each client receives the current FL model and uses it in its respective control task while collecting data.
Locally, the client updates its neural FF controller and transmits only the model updates to the central server. 
The server then aggregates these updates into a new global model using state-of-the-art federated aggregation methods. 
Then, the new global model is again distributed to the clients for inference and further local training. 
This iterative process 
\begin{itemize}
    \item \textbf{improves} control performance through the FL-based neural FF controller, 
    \item \textbf{enhances} generalization across different systems via exposure to heterogeneous local data,
    \item \textbf{reduces} communication demands by transmitting only model updates, and
    \item \textbf{addresses} data privacy concerns, by keeping raw data at the local client.
\end{itemize}
The training of the FL-based neural FF controller is summarized in Algorithm~\ref{alg:FL-based_FF}. 
Here, $G$ is the number of global communication rounds, $E$ is the number of local epochs, and $\mathcal{P}$ is the set of available clients.
In ClientTask, a closed-loop run of the client's control task using a suitable FB controller and the FL-based neural FF controller is done.
\RestyleAlgo{ruled}
\begin{algorithm}
\caption{FL-based neural FF controller.}\label{alg:FL-based_FF}
\SetAlgoLined
\textbf{Input:} Initial NN parameter $\theta^{(0)}$, ClientTask, $G$, $E$; \\
\For{$g = 0, 1, \ldots, G$} {
    Sample a subset $S_g$ of clients from set $\mathcal{P}$\;
    \For{client $i \in S_g$ in parallel} {
        Initialize local neural FF: $f_\theta^i \leftarrow \theta^{(g)}$\;
        Perform ClientTask using neural FF $f_\theta^i$\;
        Perform local optimization on data from ClientTask for $E$ local epochs to obtain local neural FF $f_{\theta'}^i$\;
    }
    Send local neural FF $f_{\theta'}^i$ to central server\;
    Update global model: $\theta^{(g+1)} = \textrm{FedAvg}\footnotemark{}(f_{\theta'}^1, \dots, f_{\theta'}^{S_g})$;
}
\end{algorithm}
\footnotetext{See~\cite{mcmahan2017communication} for details on FedAvg.}
We choose to use the state-of-the-art algorithm FedAvg, see~\cite{mcmahan2017communication}, as the FL algorithm in Algorithm~\ref{alg:FL-based_FF}. 
Nevertheless, our FL-based neural FF controller can also be used with other FL algorithms such as \textit{FedProx}~\cite{li2020_FedProx}. 
This can be especially helpful if there is significant heterogeneity in the clients performing the control task. 

\section{Control Application} \label{sec:Experimental-design}
In this section, we apply the approach presented in Section~\ref{sec:FL-and-FF-control} to a specific control task. 
This demonstrates that we can learn a neural FF controller in a multi-agent setting by using FL without sharing private data.
The proposed methods can be applied to the full range of control tasks suitable for data-driven feedforward control. 
The control task under investigation is the lateral control of an autonomous vehicle.
In these experiments various cars/clients track different trajectories in simulation, see Fig.~\ref{fig:all_paths}, and learn a joint FL-based neural FF controller for the steering input $\delta_f$ without sharing any private data, e.g., the car's longitudinal or lateral velocity, the steering input, etc.

This section is structured as follows: 
In Section~\ref{subsec:exp_background}, we provide details on the simulation environment (dynamic system, feedback controller, client paths) and the centralized neural FF controller.
Thereafter, we describe the proof-of-concept for the FL-based neural FF controller in Section~\ref{subsec:exp_PoC}.
Finally, we conclude this section by comparing the FL-based neural FF controller with the local neural FF controller for each client based solely on its local data, see Section~\ref{subsec:exp_comp_to_local}.

\subsection{Dynamic System, Feedback Controller, Client Paths \& Centralized Neural FF Controller} \label{subsec:exp_background}

\subsubsection{Dynamic Bicycle Model with Nonlinear Tires}  \label{sec:DBM}

For trajectory tracking simulations, we employ a dynamic bicycle model (DBM) that includes nonlinear tire forces, wheel-speed dynamics, and dynamic load transfer, providing a high-fidelity yet computationally efficient representation of vehicle motion~\cite{althoff2014_online_verification_automated_road_vehicles,gurtner2024role}.  
The vehicle state is defined as
\begin{equation}
    x \triangleq \left[ p_x,\, p_y,\, \psi,\, v_\mathrm{lon},\, v_\mathrm{lat},\, v_\psi,\, \omega_f,\, \omega_r\, \right]^\top,
\end{equation}
where $p_x,\,p_y$ denote global positions, $\psi$ is the yaw angle, $v_\mathrm{lon},\,v_\mathrm{lat}$ are body-fixed longitudinal/lateral velocities, $v_\psi$ is the yaw rate, and $\omega_f,\,\omega_r$ are front and rear wheel angular velocities.  
The system input $u$ is defined as
\begin{equation}
    u \triangleq \left[\delta_f,\, T_f^{\mathrm d},\, T_r^{\mathrm d} \right]^\top,
\end{equation}
with the steering angle $\delta_f$ and the demanded front and rear wheel torques $T_{f,r}^d$.  
The kinematics are
\begin{align} \label{eq:kinematics}
    \dot p_x &= v_\mathrm{lon}\cos\psi - v_\mathrm{lat}\sin\psi, \\
    \dot p_y &= v_\mathrm{lon}\sin\psi + v_\mathrm{lat}\cos\psi, \\
    \dot \psi &= v_\psi,
\end{align}
and the chassis dynamics are
\begin{align}
\dot v_{\mathrm{lon}} &= \frac{1}{m}\!\left(F_{f,\mathrm{lon}}\cos\delta_f - F_{f,\mathrm{lat}}\sin\delta_f + F_{r,\mathrm{lon}}\right) + v_{\mathrm{lat}} v_\psi,
\label{eq:dynamics_v_long}\\
\dot v_{\mathrm{lat}} &= \frac{1}{m}\!\left(F_{f,\mathrm{lon}}\sin\delta_f + F_{f,\mathrm{lat}}\cos\delta_f + F_{r,\mathrm{lat}}\right) - v_{\mathrm{lon}} v_\psi,
\label{eq:dynamics_v_lat}\\
I_z \dot v_\psi &= l_f\!\left(F_{f,\mathrm{lon}}\sin\delta_f + F_{f,\mathrm{lat}}\cos\delta_f\right) - l_r F_{r,\mathrm{lat}},
\label{eq:dynamics_v_psi}
\end{align}
with vehicle mass $m$, yaw moment of inertia $I_z$, and front/rear distances to the center of gravity (CoG) $l_f$ and $l_r$.
Longitudinal and lateral forces for the front and rear wheels, $F_{i,\mathrm{lon}}$ and $F_{i,\mathrm{lat}}$, $i=\{f,r\}$, are modeled using Pacejka’s Magic Formula~~\cite{pacejka1992magic}
and constrained via the friction ellipse, considering the street friction coefficient $\mu$, see Tab.~\ref{tab:magic}.  
The wheel speeds are governed by
\begin{align}
    I_w \dot\omega_f &= T_{f}^d - R F_{f, \mathrm{lon}} - M_{\mathrm{rr},f}, \\
    I_w \dot\omega_r &= T_{r}^d - R F_{r, \mathrm{lon}} - M_{\mathrm{rr},r},
\end{align}
with wheel radius $R$ and the respective rolling-resistance torque $M_{\mathrm{rr, i}}=R F_{z,i} c_{\mathrm{rr}}\,\mathrm{sgn}(\omega_{i})$ with normal forces $F_{z,i}$ for $i \in \{f,r\}$\footnote{For numerical stability, the wheel dynamics are integrated with $N_{\mathrm{sub}}$ explicit Euler substeps per main step $\Delta t$.}.
Simulation parameters are given in Table~\ref{tab:system_parameter}.
\begin{table}
    \centering
    \caption{Simulation parameters.}
    \label{tab:system_parameter}
    \begin{tabular}{cccc}
        \toprule
        \textbf{Symbol} &\textbf{Meaning}    & \textbf{Unit}         &  \textbf{Value}   \\ \hline
        \midrule
        $m$ & mass & \si{\kilogram} & $2273$ \\
        $I_z$ & yaw moment of inertia & \si{\kilogram\meter\squared} & $4423$ \\
        $I_w$ & wheel moment of inertia & \si{\kilogram\meter\squared} & $1.5$ \\
        $R$ & wheel radius & \si{\meter} & $0.35$ \\
        $l_f$ & CoG to front axle & \si{\meter} & $1.292$ \\
        $l_r$ & CoG to rear axle & \si{\meter} & $1.515$ \\
        $\mu$ & street friction coefficient & -- & $1.0$ \\
        $c_{\mathrm{rr}}$ & rolling friction coefficient & -- & $0.01$ \\
        $\Delta t$ & main time step & \si{\second} & $0.05$ \\
        $N_{\mathrm{sub}}$ & Euler substeps & -- & $10$ \\
        \bottomrule
        \end{tabular}
\end{table}
\begin{table}[t]
\centering
    \caption{Magic-Formula coefficients used for the front and rear axle, as well as the longitudinal and lateral direction.}
    \label{tab:magic}
    \begin{tabular}{ccccc}
        \toprule
        \textbf{Set} & \textbf{B} & \textbf{C} & \textbf{D} & \textbf{E} \\ \hline
        \midrule
        Front lateral & $9.3$ & $1.1$ & $1.0$ & $-0.25$ \\
        Rear lateral & $10.0$ & $1.2$ & $1.0$ & $-0.25$ \\
        Front longitudinal & $9.3$ & $1.3$ & $1.0$ & $-0.25$ \\
        Rear longitudinal & $10.0$ & $1.3$ & $1.0$ & $-0.25$ \\
        \bottomrule
    \end{tabular}
\end{table}

\subsubsection{Client Paths}
\begin{figure}[h!]
    \centering
    \includegraphics[width=0.99\linewidth]{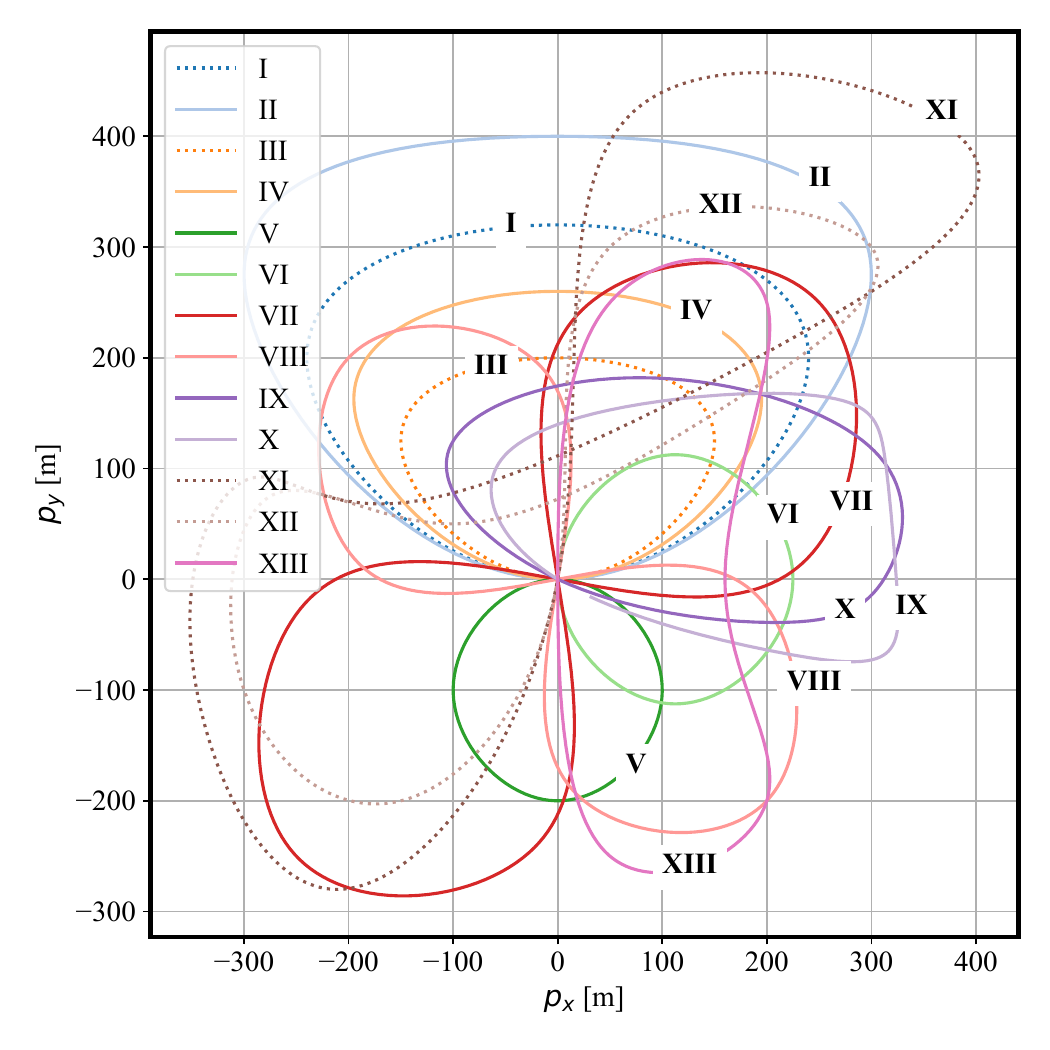}
    \caption{Paths in world coordinates. Train paths are indicated by the solid line, test paths by the dotted line (I, III, XI, XII).}
    \label{fig:all_paths}
\end{figure}

\begin{figure}[t]
    \centering
    \includegraphics[width=0.99\linewidth]{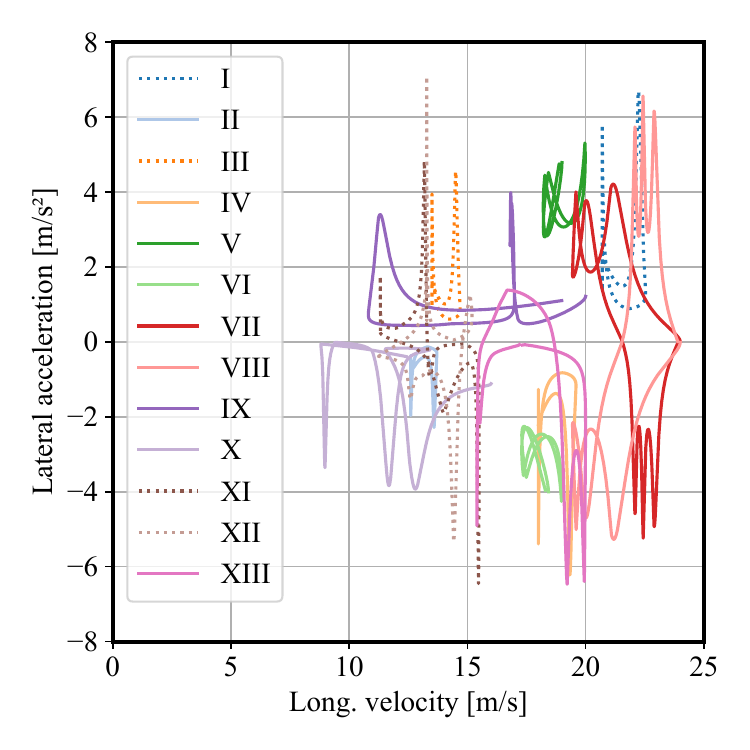}%
    \caption{Desired operating conditions across trajectories: desired longitudinal velocity versus desired lateral acceleration for Paths I–XIII. 
    The curve length is proportional to the number of trajectory samples. 
    Solid lines indicate training paths, dotted lines indicate test paths. 
    }
    \label{fig:all_paths_lat_acc}
\end{figure}

In order to evaluate the closed-loop performance and to provide representative training data for the FL-based neural feedforward controller, a diverse set of reference paths was generated. 
The set comprises 13 paths (Path I-XIII) divided into training (solid) and test (dotted) sets. 
Figure~\ref{fig:all_paths} illustrates the trajectories in world coordinates $p_x, p_y$, covering a wide range of curvature profiles and geometric complexities. 
Figure~\ref{fig:all_paths_lat_acc} depicts the corresponding desired longitudinal velocities and lateral acceleration.
Several maneuvers involve substantial lateral dynamics with peak accelerations above \SI{6}{m/s^2}. 
The designed paths challenge both the stabilizing feedback controller and the learned feedforward component.
Thus, they enable a meaningful assessment of tracking performance under demanding conditions.

\subsubsection{Feedback Controller} \label{subsubsec:FB}
We use an adaptation of the FB controller presented in~\cite{althoff2014_online_verification_automated_road_vehicles} with a reference frame moving along the reference trajectory.
%
%
The lateral tracking error $\varepsilon_y$ and longitudinal tracking error $\varepsilon_x$ are defined as
\begin{align}
    \begin{aligned}
        \varepsilon_{x} &= \cos\psi^d (p_x^d - p_x^a) + \sin\psi^d  (p_y^d - p_y^a), \\
        \varepsilon_{y} &= -\sin\psi^d (p_x^d - p_x^a) + \cos\psi^d (p_y^d - p_y^a),
    \end{aligned} \label{eq:errors}
\end{align}
where superscripts $d$ and $a$ mark the desired and actual values, respectively.
The FB control signal for the steering input is given by 
\begin{align} \label{eq:FB_steer}
     u_{\delta}^{\mathrm{FB}} = K_1 \varepsilon_y + K_2 (\psi^d - \psi^a) + K_3 (v_\psi^d - v_\psi^a),
\end{align}
combining the lateral tracking error, the yaw angle error, and the yaw rate error using controller gains $K_1$, $K_2$, and $K_3$ as specified in Tab.~\ref{tab:ctrl_gains}.

The combined steering input $u_{\delta} = u_{\delta}^{\mathrm{FB}} + u_{\delta}^{\mathrm{FF}}$, i.e., the desired front wheel angle $\delta_f^d$, is further affected by an underlying steering rate controller using proportional control with gain $K_4$ based on the difference between the desired and actual front wheel angles as
\begin{align} \label{eq:steering-rate-controller}
    \dot \delta_f = K_4 (\delta_f^d - \delta_f^a),
\end{align}
see~\cite{althoff2014_online_verification_automated_road_vehicles} for more details.
\begin{table}[t]
    \centering
    \caption{Controller gains used for the feedback controller.}
    \begin{tabular}{clc}
        \toprule
        \textbf{Gains} & \textbf{Meaning} & \textbf{Value}  \\ \hline
        \midrule
        $K_1$ & Lateral error & 0.5 \\
        $K_2$ & Orientation error & 3  \\
        $K_3$ & Yaw rate error & 1 \\
        $K_4$ & Steering velocity & 2 \\
        $K_5$ & Long. error & 1  \\
        $K_6$ & Long. velocity error & 5 \\
        \bottomrule
    \end{tabular}
    \label{tab:ctrl_gains}
\end{table}
The feedback controller for the longitudinal velocity is given by
\begin{align} \label{eq:FB_v}
    a_\mathrm{lon}^{\mathrm{FB}} = K_5 \varepsilon_x + K_6 (v_\mathrm{lon}^d - v_\mathrm{lon}^a)
\end{align}
with gains $K_5$ and $K_6$~\cite{althoff2014_online_verification_automated_road_vehicles}.
The desired longitudinal acceleration $a_\mathrm{lon}^{\mathrm{FB}}$ is converted into front- and rear-wheel torque commands. 
A torque distribution factor $\tau_{\mathrm{f}}$ assigns the proportion between front and rear axles. 
It is set to $\tau_{\mathrm{f,accel}} = 1$ during acceleration and $\tau_{\mathrm{f,decel}} = 0.75$ during braking. 
The resulting torques are computed as
\begin{align}
    T^d_f &= a_\mathrm{lon}^{\mathrm{FB}}\, m R \tau_{\mathrm{f}},  \\
    T^d_r &= a_\mathrm{lon}^{\mathrm{FB}}\, m R (1 - \tau_{\mathrm{f}}),
\end{align}
ensuring a physically consistent force distribution.

\begin{figure}[h!]
    \centering
    \includegraphics[width=0.99\linewidth]{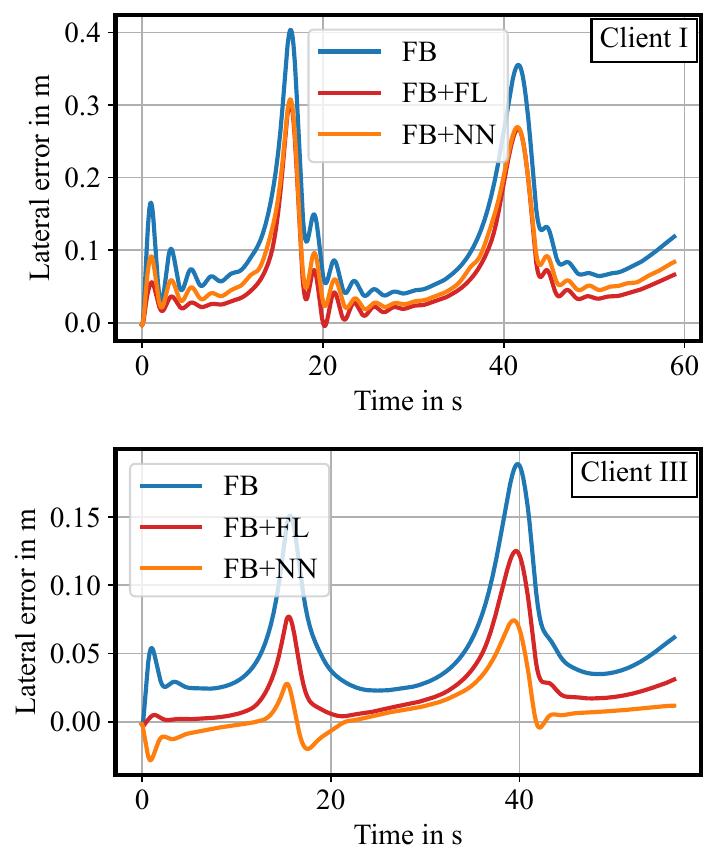}
    \caption{Lateral errors for clients I and III with FB control (blue), federated neural FF control (red), and centralized neural FF control (orange) after $G=5$ global rounds and $E=3$ local epochs (both combined with FB).
    }
    \label{fig:testpaths-lateral-error_1}
\end{figure}

\begin{figure}[h!]
    \centering
    \includegraphics[width=0.99\linewidth]{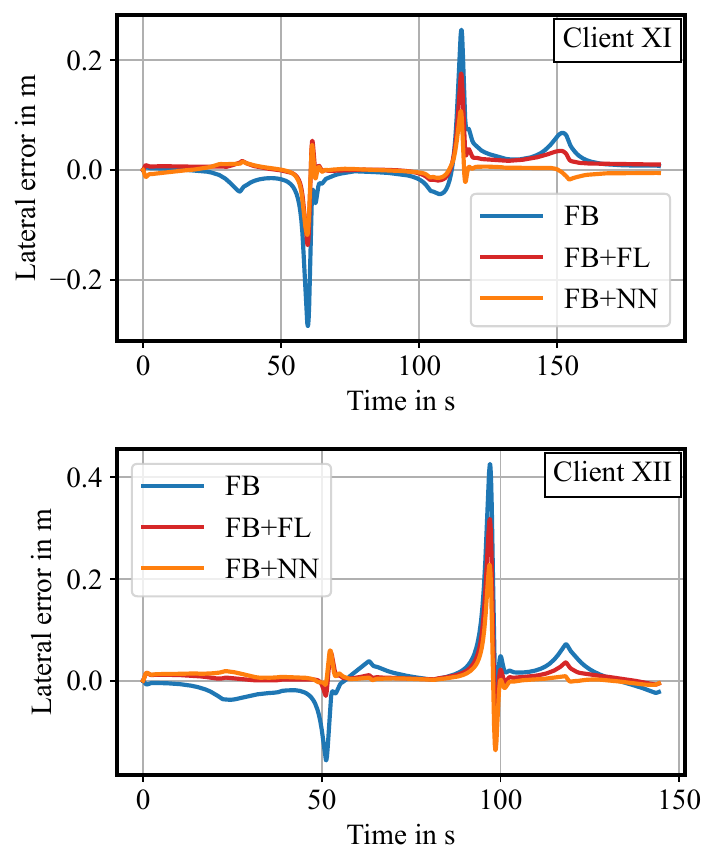}
    \caption{Lateral errors for clients XI and XII with FB control (blue), federated neural FF control (red), and centralized neural FF control (orange) after $G=10$ global rounds and $E=3$ local epochs (both combined with FB).
    }
    \label{fig:testpaths-lateral-error_2}
\end{figure}

To evaluate the performance of the control system, we use the mean and maximum lateral tracking error (LTE) along a trajectory 
\begin{align} 
    \text{LTE}_{\mathrm{mean}} &=\frac{1}{T} \int_0^T |\varepsilon_{y}(t)|  \ \mathrm{d}t, \label{eq:mean_LTE}\\
    \text{LTE}_\mathrm{max} &= \max_t |\varepsilon_{y}(t)| , \label{eq:max_LTE}    
\end{align}
with the trajectory duration $T$.
We approximate~\eqref{eq:mean_LTE} as a finite sum with the corresponding time step $\Delta t=0.05$.
Using this feedback controller, we have already achieved reasonable trajectory tracking on the test trajectories we selected (see Figs.~\ref{fig:testpaths-lateral-error_1} and~\ref{fig:testpaths-lateral-error_2}).

\subsubsection{Centralized Neural Feedforward Controller} \label{subsec:centralized_ff}
In this section, we learn a \emph{preview–based} feedforward (FF) controller, guided by the system’s discrete-time \emph{relative degree}, to improve lateral tracking. 
For the considered vehicle model, the lateral dynamics with the yaw-rate output $v_\psi$ have relative degree $r=2$ with respect to the steering input $\delta_f$. 
Hence, a preview–based FF law at time $k$ may depend on a future desired state at $k{+}r$.

\noindent \textit{Feature selection:}
The dynamic bicycle model in \eqref{eq:dynamics_v_long}–\eqref{eq:dynamics_v_psi} shows that the lateral subsystem depends on the by longitudinal speed $v_{\mathrm{lon}}$, yaw rate $v_\psi$, and steering $\delta_f$~\cite{velenis2010steady}.
The kinematic pose $(p_x,p_y,\psi)$ does not causally influence the local motion dynamics. 
We therefore use $(v_{\mathrm{lon}}, v_\psi)$ as physics-motivated features and exploit the relative-degree preview. 
The neural FF model takes the current values and an $r$-step preview as inputs
\begin{equation} \label{eq:NN}
    \mathrm{NN}\!\big(v_{\mathrm{lon}}[k],\, v_\psi[k],\, v_{\mathrm{lon}}[k{+}r],\, v_\psi[k{+}r]\big).
\end{equation}

\noindent \textit{Training with a shifted horizon:}
We train the network on recorded closed-loop trajectories using a shifted horizon to respect causality. 
For each time $k$, we form the input as in~\eqref{eq:NN} with measured quantities and learn the applied steering $\delta_f[k]$ using the mean-squared loss
\begin{equation} \label{eq:ff_mse}
    \mathcal{L}_{\mathrm{MSE}} =\frac{1}{|\mathcal{D}|}\sum_{k\in\mathcal{D}} \!\left(\delta_f[k] -\mathrm{NN}(*) \right)^{2},
\end{equation}
with $* = \left[ v^a_{\mathrm{lon}}[k], v^a_\psi[k], v^a_{\mathrm{lon}}[k{+}r], v^a_\psi[k{+}r] \right]$.
At run time, the FF controller uses only desired trajectories
\begin{equation*}
    u_{\mathrm{FF}}[k] = \mathrm{NN}\!\big(v_{\mathrm{lon}}^{d}[k],\, v_\psi^{d}[k],\, v_{\mathrm{lon}}^{d}[k{+}r],\, v_\psi^{d}[k{+}r]\big)
\end{equation*}
and is combined with the stabilizing feedback term as $\delta_f^d[k]=u_{\mathrm{FB}}[k]+u_{\mathrm{FF}}[k]$. 
This \emph{train-with-shift, evaluate-with-preview} setup preserves causality and adheres to the FF principle while remaining straightforward to learn from closed-loop trajectory data.

\noindent \textit{Architecture and regularization:}
We use a fully connected network with three hidden layers of width $n_{\mathrm{neurons}}$ and ReLU activations. 
To improve robustness, all layers are spectrally normalized, which bounds the Lipschitz constant and mitigates unstable learning dynamics; see also~\cite{shi2019_neural_lander}. 
The network parameters are optimized with Adam~\cite{kingma2017adam} (learning rate $\eta=0.001$, batch size $B=32$). 
Unless stated otherwise, the same hyperparameters are used in the federated and local settings. 
Based on preliminary tuning, $n_{\mathrm{neurons}}=5$ (81 trainable parameters) proved sufficient to learn the FF mapping.
\subsection{FL-based Neural FF Controller} \label{subsec:exp_PoC}
In the proposed learning setting, a set of identical clients collaboratively learn a shared neural FF controller, each using locally collected trajectory data.
Although the clients share identical models, they follow unequal trajectories, resulting in heterogeneous datasets.
This effectively forms a distributed design-of-experiment setup.
The test clients I, III, XI, and XII (dotted in Fig.~\ref{fig:all_paths}) are used exclusively for evaluation. 
This setup is representative of many IoT or fleet-level learning scenarios, where similar devices operate under diverse conditions and thus generate complementary, non-overlapping datasets.
Specifically, it corresponds to an in-fleet learning setting, in which multiple instances of the same vehicle model collaboratively optimize a shared controller through federated updates derived from their respective local experiences.

\begin{figure}[!ht]
  \centering
  \includegraphics[width=0.99\linewidth]{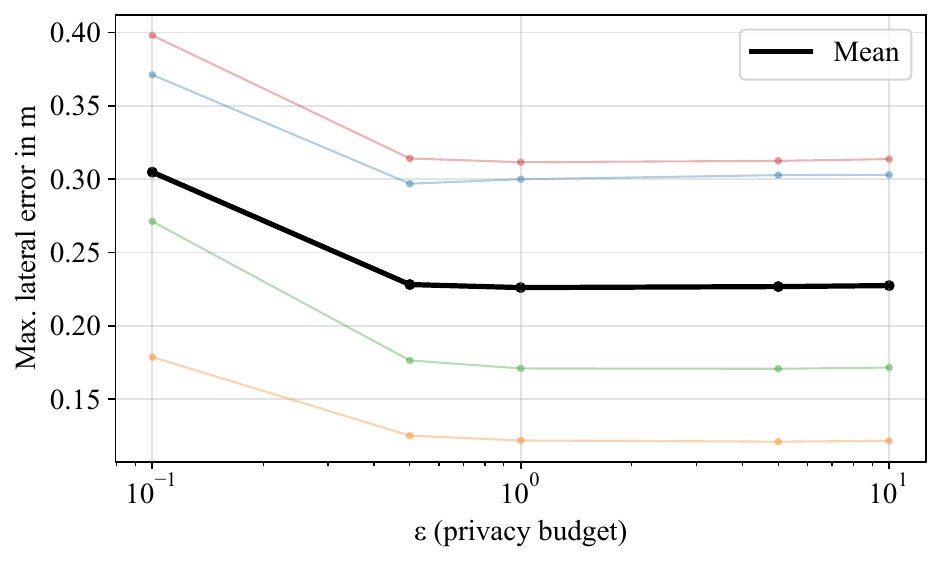}
  \caption{Maximum absolute lateral error vs.\ privacy budget $\varepsilon$ (log scale) for $G=10$ global communication rounds. 
  The black line indicates the mean over the four test paths (I, III, XI, and XII.
  Performance varies only marginally over $\varepsilon\in[0.1,10]$. 
  We therefore use $\varepsilon=1$ with $\delta=10^{-5}$ in the main experiments.}
  \label{fig:dp-privacy-budget}
\end{figure}

The training procedure follows the standard FedAvg protocol~\cite{mcmahan2017communication}:
\begin{itemize}
    \item Each client receives the global model, trains locally on its trajectory data, and sends the parameter gradients to the server.
    \item The server aggregates all updates into a new global model.
\end{itemize}
\noindent \textit{Differential Privacy:}
To increase privacy, we enforce \emph{example-level} differential privacy during each client’s local training via DP-SGD~\cite{abadi2016deep}: per-example gradients are clipped to an $\ell_2$ norm cap of $C$, and Gaussian noise with standard deviation $\sigma C$ is added to the clipped mean before performing the optimizer step.
For hyperparameter tuning, we sweep the privacy budget $\varepsilon \in [0.1,10]$ and track privacy loss with a Rényi DP accountant~\cite{mironov2017renyi} (fixed $\bar\delta=10^{-5}$) using \texttt{Opacus}~\cite{opacus}. 
As shown in Fig.~\ref{fig:dp-privacy-budget}, the maximum lateral error is largely insensitive to the privacy budget $\varepsilon$.
Hence, we adopt $\varepsilon=1$ as our default. 
Unless stated otherwise, we use batch size $B=32$, clipping norm $C=0.5$, and $3$ local epochs per round.\footnote{Privacy unit: a single training \emph{example} on a client.}

\subsection{Comparison to Local Neural FF} \label{subsec:exp_comp_to_local}
Finally, we benchmark the FL-based neural FF controller against purely local training.
Herein, each training client learns its own neural FF controller using only local data.
This setting does not allow communication between clients. 
Therefore, knowledge cannot be shared across the network.

For this experiment, each client completes $G=10$ training rounds exclusively on its assigned trajectory.
After each round, it updates its neural FF controller with $E=3$ local epochs. 
To test the generalization capabilities, the resulting controllers are evaluated on the test paths depicted in Fig.~\ref{fig:all_paths}.

\section{Results \& Discussion} \label{sec:Results-and-discussion}
In this section, we present and analyze the results of the experiments elaborated in Section~\ref{sec:Experimental-design}.
The closed-loop tracking performance is quantified using the mean~\eqref{eq:mean_LTE} and maximum~\eqref{eq:max_LTE} lateral error, respectively.
It is important to emphasize that these error measures are not explicitly optimized during the learning procedures. 
Instead, they serve as independent performance criteria and provide an evaluation of how well the learned controllers generalize in closed-loop control.

\subsection{FL-based Neural FF control} 
\begin{figure}[!h]
    \centering
    \includegraphics[width=0.99\linewidth]{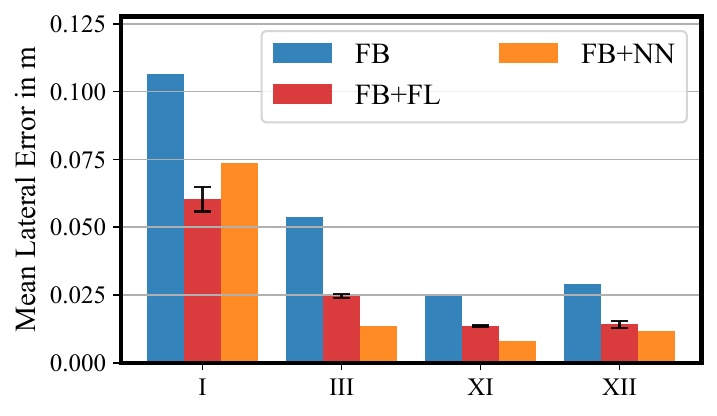}
    \caption{Mean lateral error for the FB controller (FB, blue), the FB controller with FL-based neural FF controller (FB+FL, red), and with the centralized neural FF controller (FB+NN, orange) for the test clients I, III, XI, and XII after $G=10$ global communication rounds with $E=3$ local epochs.
    The depicted values for the federated model are the mean over 20 independent initializations. 
    The corresponding standard deviation is also shown.
    }
    \label{fig:mean_LTE}
\end{figure}

\begin{figure}[!ht]
    \centering
    \includegraphics[width=0.99\linewidth]{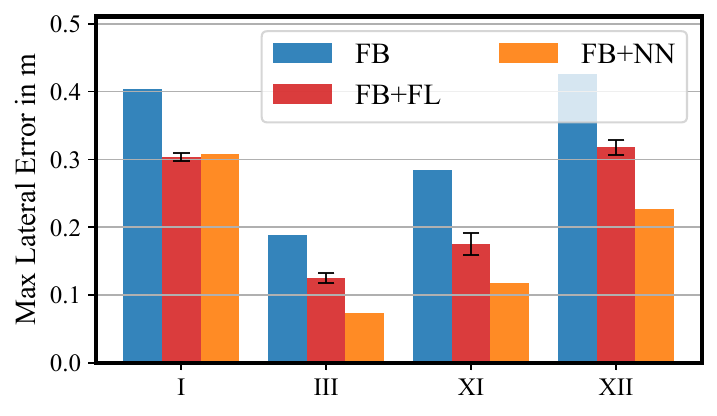}
    \caption{Maximum lateral error for the FB controller (FB, blue), the FB controller with FL-based neural FF controller (FB+FL, red), and with the centralized neural FF controller (FB+NN, orange) for the test clients I, III, XI, and XII after $G=10$ global communication rounds with $E=3$ local epochs.
    The depicted values for the federated model are the mean over 20 independent initializations.
    The corresponding standard deviation is also shown.
    }
    \label{fig:max_LTE}
\end{figure}

Figures~\ref{fig:testpaths-lateral-error_1} and~\ref{fig:testpaths-lateral-error_2} show the lateral tracking error for Clients~I, III, XI, and~XII. 
In all cases, the learning-based feedforward (FF) controllers substantially reduce deviations relative to the feedback-only baseline (FB).
The federated model (FB+FL) closely matches the centralized neural FF controller (FB+NN) on Clients~XI and~XII, and even slightly surpasses it on Client~I.
On Client~III, the FL setup still lags the centralized one.
Figures~\ref{fig:mean_LTE} and~\ref{fig:max_LTE} report the mean and maximum lateral error over 20 independent, randomly initialized neural network runs, where the error bars show the standard deviation. 
This representation leads to three key observations: 
(i) Both learned FF controllers (FB+NN and FB+FL) substantially outperform the feedback-only baseline (FB) on every client, for both mean and max error. 
(ii) The centralized model (FB+NN) attains the lowest errors overall, except for Client~I.
(iii) The federated model (FB+FL) trained with differential privacy tracks the centralized model closely on average error and is only modestly worse on maximum error.
Across all initializations, error bars for the federated model are small, indicating low run-to-run variability.

Overall, the FL-based controller effectively maintains high tracking accuracy despite decentralized training. 
The close agreement between the FL-based and centralized performance confirms that Federated Learning can achieve near-centralized performance while substantially reducing communication costs and increasing data privacy.

\subsection{Comparison to Local Neural FF}
\begin{figure*}[!ht]
    \centering
    \includegraphics[width=0.9\linewidth]{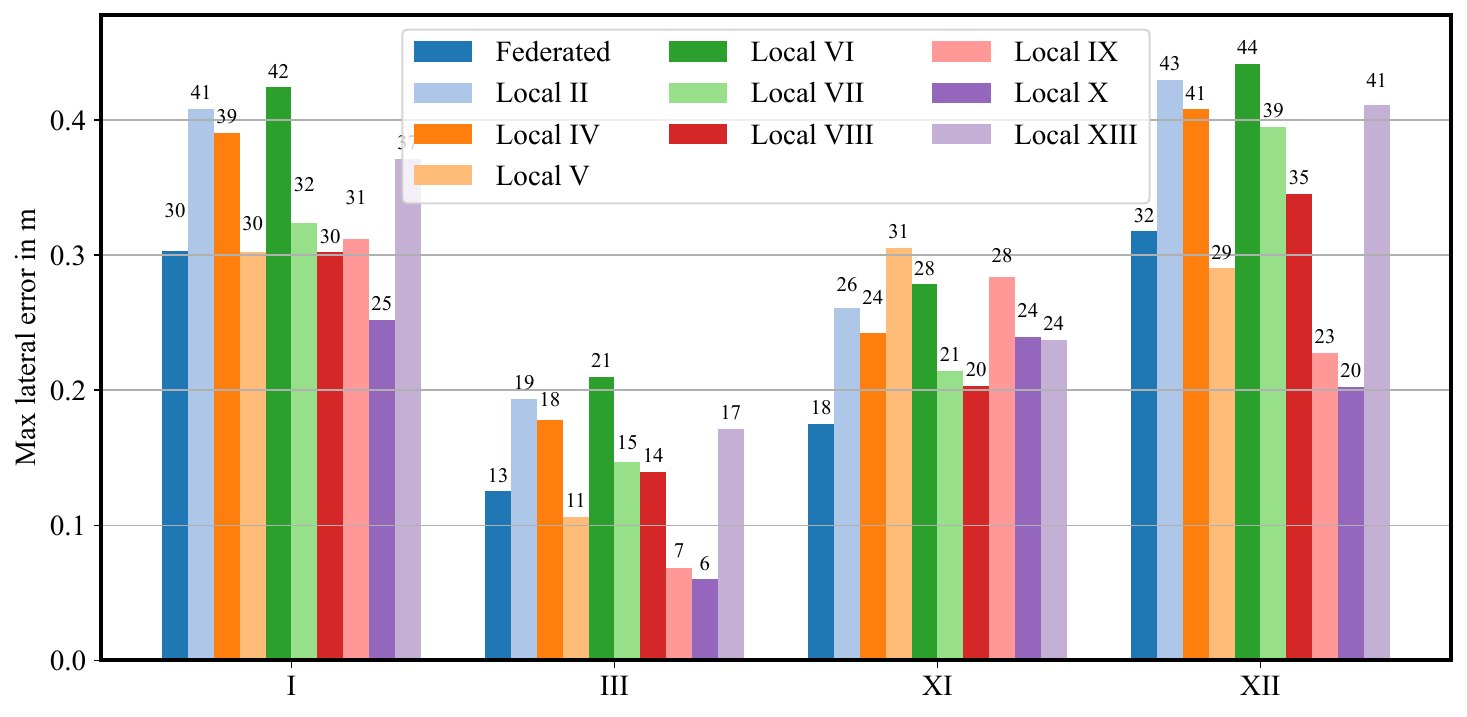}
    \caption{
    Maximum lateral tracking error~\eqref{eq:max_LTE} for the federated and locally trained models evaluated on test clients I, III, XI, and XII. 
    The federated model yields consistently low errors across clients.
    The local models show large variability, illustrating the generalization advantage of federated learning over single-client training.
    The values are depicted in $\si{\centi \meter}$. }
    \label{fig:exp_local_vs_fed_ratio}
\end{figure*}

We compare the FL-based neural FF controller to a local-only baseline, in which each client trains a controller solely on its own trajectory data.
Figure~\ref{fig:exp_local_vs_fed_ratio} shows the maximum lateral errors $\mathrm{LTE}_{\mathrm{max}}$ across all locally trained models and test paths. 

The FL-based neural FF controller generalizes well to all test clients, consistently matching or outperforming the locally trained models. 
No local model achieves consistent improvements across all test paths. 
Some local models perform better on individual test paths, such as Local X on path I and XII or Local IX on path XII.
This behavior is expected, as locally trained controllers can overfit to the data from their respective trajectories, yielding good performance only within that narrow regime.

These observations highlight the robustness of the federated approach.
By aggregating knowledge from multiple clients, FL mitigates overfitting and yields a neural FF controller that generalizes reliably across diverse operating conditions. 
Sharing information through FL thus provides an effective and scalable strategy for distributed learning of feedforward controllers.
Together with the previous results, the local-only controllers define a lower performance bound, while the centralized training represents the upper bound achievable with full data access.
The FL-based neural FF controller closes most of this gap, achieving near-centralized performance while maintaining full data decentralization and increasing privacy.

\subsection{Communication-Efficiency} 
\begin{figure}
    \centering
    \includegraphics[width=0.99\linewidth]{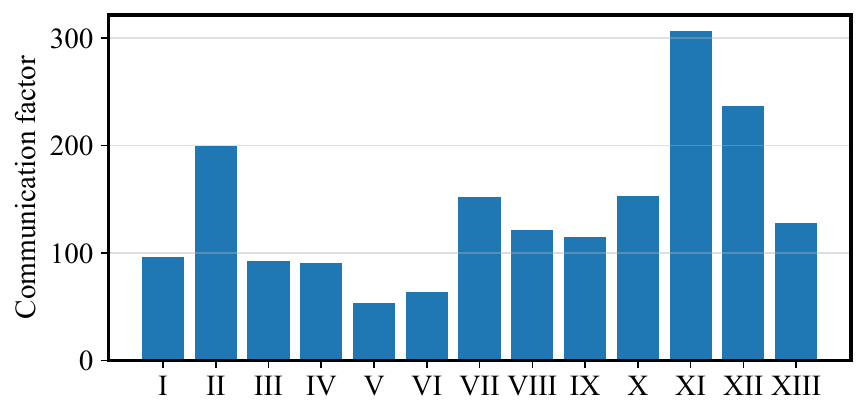}
    \caption{Communication transfer factor for all paths.}
    \label{fig:comm-transfer-factor}
\end{figure}

In a centralized setup, each client must upload its entire dataset to the server for training in each round.
Federated Learning avoids this by keeping data local: clients train their neural FF controller on-device and transmit only model parameters to the server.
The resulting reduction in uplink communication cost per round is quantified by the \emph{communication transfer factor}, defined as the ratio between local dataset size (centralized) and model size (federated).
The downlink communication cost, i.e., the updated model, is identical in both the centralized and the federated setups.
In the conducted experiments, local datasets span between \SI{30}{\second} (Client~V) and \SI{190}{\second} (Client~XI), with a sampling time of $\Delta t=\SI{0.05}{\second}$. 
Figure~\ref{fig:comm-transfer-factor} shows that the communication transfer factor exceeds~50 for short trajectories (e.g., Path~V) and surpasses~200 for long or complex ones (e.g., Paths~II).
Thus, transmitting raw data would require up to two orders of magnitude more bandwidth than sending model updates, see Fig.~\ref{fig:comm-transfer-factor}.
These results clearly demonstrate that FL can drastically reduce network and central server load while maintaining learning capability.
By exchanging compact model updates rather than full datasets, FL enables scalable and privacy-aware deployment across large IoT or fleet control systems.
\section{Conclusion \& Outlook} \label{sec:Conclusion-and-outlook}
%
To address key challenges of communication efficiency and privacy preservation in multi-agent feedforward control, this work introduces a novel approach based on Federated Learning (FL).
The proposed method eliminates the need for raw data sharing while still achieving accurate and robust control performance. 
The FL-based controller iteratively aggregates local model updates into a global model, thereby improving trajectory tracking performance across multiple clients.
Additionally, differential privacy is applied to local training to improve privacy-preservation.

Simulation results on vehicle trajectory tracking demonstrated that the FL-based neural FF controller achieves performance comparable to centralized FF training, despite requiring significantly less communication while maintaining data privacy. 
Compared to local-only learning, the federated approach further improves generalization, confirming the benefits of information sharing across clients even in a decentralized setting.

Future work will extend this proof-of-concept in several directions. 
First, advanced FL techniques such as clustered FL will be explored to explicitly handle client heterogeneity and more complex vehicle dynamics. 
Second, experiments with real-world platforms will provide further validation under practical conditions. 
Finally, optimizing communication protocols and ensuring robustness against faulty or non-cooperative clients are critical steps toward scalable and reliable deployment of FL-based feedforward control in control systems.

\bibliographystyle{IEEEtran}
\bibliography{bibliography}

\end{document}